\documentclass{article} 
\usepackage{nfam2026_workshop}
\usepackage{times}
\usepackage{graphicx} 
\usepackage{amsmath}
\usepackage{amssymb}

\usepackage{amsmath,amsfonts,bm}

\def\eqref#1{equation~\ref{#1}}

\def\1{\bm{1}}

\DeclareMathAlphabet{\mathsfit}{\encodingdefault}{\sfdefault}{m}{sl}
\SetMathAlphabet{\mathsfit}{bold}{\encodingdefault}{\sfdefault}{bx}{n}

\usepackage{hyperref}
\usepackage{url}

\title{Reasoning as Attractor Dynamics: Latent Memory Retrieval via Gibbs-Weighted Energy Minimization}

\author{Kanishk Awadhiya\\
Independent Researcher\\
\texttt{kanishk.awadhiya@gmail.com}
}

\iclrfinalcopy 
\begin{document}

\maketitle

\begin{abstract}
Large Language Models (LLMs) are traditionally viewed as autoregressive generators. However, from the perspective of collective computation, they function as high-dimensional Dense Associative Memories that store complex reasoning patterns as latent attractors. In this work, we investigate the energy landscape of mathematical reasoning. We posit that correct reasoning chains correspond to deep, wide attractor basins (``flat minima'') in the model's output distribution, whereas hallucinations manifest as sharp, unstable local minima. To exploit this geometry, we introduce a retrieval mechanism based on a Gibbs measure of the trajectory's spectral entropy. By sampling multiple reasoning paths and weighting them by their inverse energy ($P \propto e^{-\beta E}$), we approximate the equilibrium distribution of the associative memory, effectively ``relaxing'' the system into a robust solution. Empirically, this physics-inspired mechanism improves Microsoft Phi-3.5 performance on GSM8K by 5.38\% (84.7\% $\to$ 90.1\%), demonstrating that inference is better modeled as a dynamic settling process into an attractor basin rather than greedy next-token prediction.
\end{abstract}

\section{Introduction}

The recent resurgence of Associative Memory (AM) has provided a unifying framework linking Energy-Based Models (EBMs), Hopfield Networks, and Transformers \citep{krotov2016dense, ramsauer2020hopfield, hoover2024energy}. While Transformers are typically trained via maximum likelihood, their inference dynamics can be understood as querying a content-addressable memory: the input prompt cues a retrieval process that reconstructs a stored pattern (the completion).

However, a fundamental disconnect exists between the storage capacity of these models and their retrieval dynamics. Standard decoding algorithms---Greedy Search, Nucleus Sampling---treat generation as a kinetic process that often gets trapped in local minima. In reasoning tasks, these local minima manifest as ``confident hallucinations'': answers that have high local probability (low energy) but lack structural stability. We propose a shift in perspective: \textit{Reasoning is not just generation; it is memory retrieval via attractor dynamics.}

In this paper, we operationalize this view by treating generated reasoning paths as particles in an energy landscape. We define the \textit{Trajectory Energy} $E(y)$ as the spectral entropy (sequence NLL) of the path. Drawing on the principles of thermodynamics and statistical mechanics \citep{lecun2006tutorial}, we argue that correct solutions reside in \textit{Flat Minima}---regions of high volume and low curvature \citep{garipov2018loss}.

We introduce \textbf{Gibbs-Weighted Basin Selection}, a test-time mechanism that:
\begin{enumerate}
    \item Explores the landscape via high-temperature sampling (creating a particle cloud).
    \item Evaluates the stability of each particle via its spectral entropy.
    \item Relaxes the system into the global minimum using a Gibbs distribution ($W \propto E^{-2}$).
\end{enumerate}

Our approach bridges the gap between Modern Hopfield Networks (which minimize energy) and LLM Reasoning, showing that a physics-based retrieval operator can extract SOTA performance (90\% on GSM8K) from a small 3.8B parameter model \citep{microsoft2024phi3}.

\section{Theoretical Framework}

\subsection{The Transformer as an Energy-Based Model}
Following the seminal work on Dense Associative Memories \citep{krotov2016dense} and Energy Transformers \citep{hoover2024energy}, we can view the probability assigned by an LLM to a sequence $y$ given input $x$ as a Boltzmann distribution:
\begin{equation}
P_\theta(y|x) = \frac{e^{-E_\theta(y, x)}}{Z(x)}
\end{equation}
where $E_\theta(y, x)$ is the energy function implicitly defined by the network's weights, and $Z(x)$ is the intractable partition function. In standard training, we minimize the negative log-likelihood (NLL), which is equivalent to minimizing the energy of the data samples.

\begin{figure}[t]
\begin{center}
\includegraphics[width=0.8\linewidth]{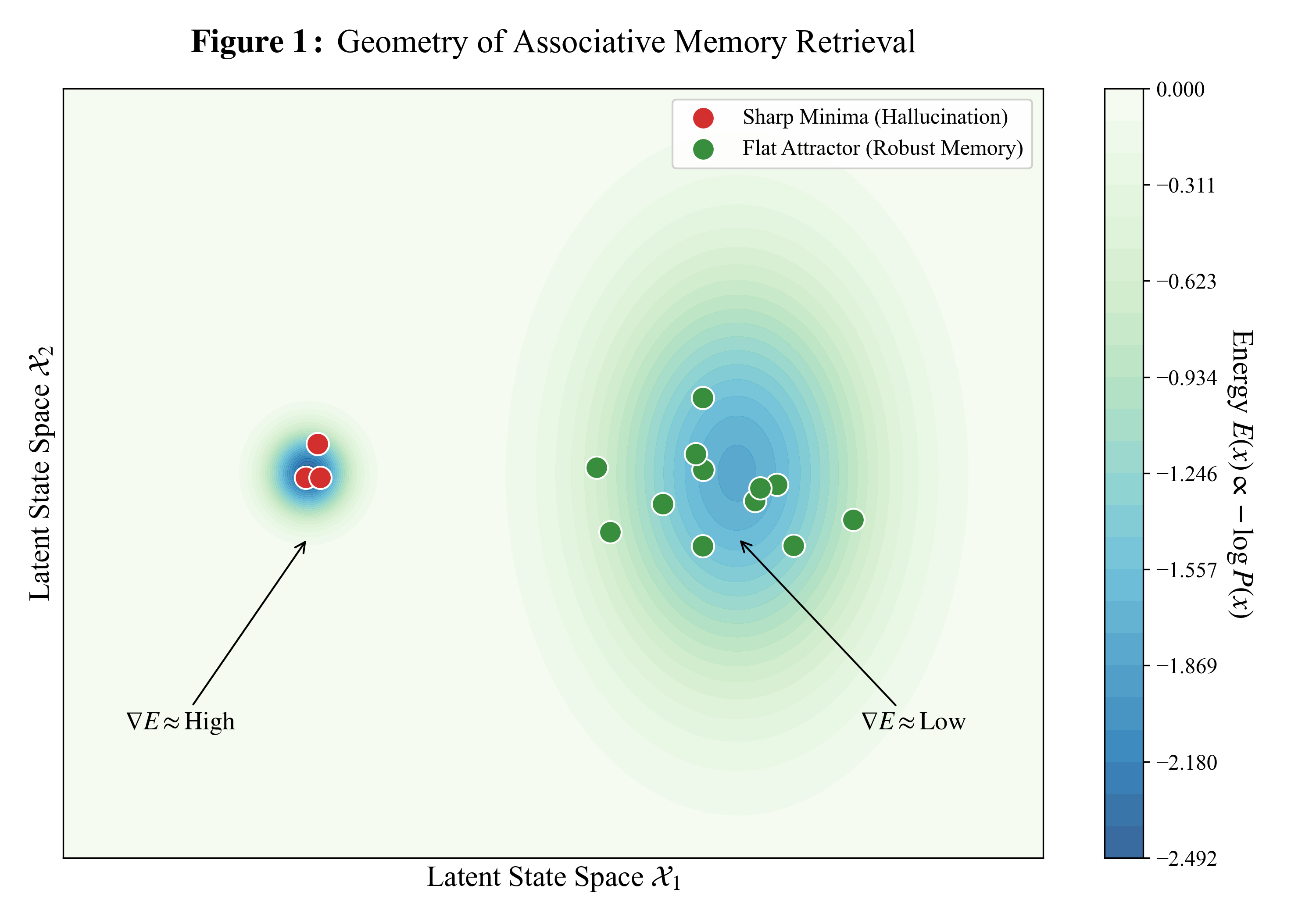}
\end{center}
\caption{Geometry of Associative Memory Retrieval. Correct reasoning chains correspond to \textit{Flat Minima} (Attractor Basins) with high entropic volume, while hallucinations are often \textit{Sharp Minima} (Metastable States). Standard greedy decoding can get trapped in sharp minima, while our Gibbs-Weighted mechanism allows the system to relax into the robust attractor.}
\label{fig:landscape}
\end{figure}

\subsection{Energy Landscapes: Sharp vs. Flat Minima}
A core tenet of our work is the geometric distinction between robust memories and hallucinations (Figure \ref{fig:landscape}).

\textbf{Sharp Minima (Hallucinations):} These are narrow valleys in the energy landscape. While the energy $E(y)$ might be low (high probability), the surrounding volume is small. This implies the solution is brittle; a slight perturbation in the generation path (or ``noise'' in the memory query) leads to a completely different, often incorrect, state.

\textbf{Flat Minima (Robust Memories):} Generalizable solutions tend to lie in ``flat minima''---regions where the energy surface is convex and wide \citep{garipov2018loss}. In an associative memory, this corresponds to a \textit{Basin of Attraction} with a large basin of entropic volume. Even if the global minimum of the sharp peak is lower, the probability mass (integral of volume) of the flat basin is higher.

\subsection{Retrieval as Gibbs Sampling}
Standard decoding resembles finding the mode $\arg\min_y E(y)$. However, in a rugged landscape, the global minimum is hard to find greedily. We propose a stochastic retrieval process. We sample a set of trajectories $\mathcal{Y} = \{y^{(1)}, \dots, y^{(K)}\}$. We then re-weight these trajectories using a post-hoc Gibbs measure:
\begin{equation}
P_{\text{retrieval}}(y^{(k)}) \propto \exp\left(-\beta \cdot \mathcal{H}(y^{(k)})\right)
\end{equation}
where $\mathcal{H}(y^{(k)})$ is the Spectral Entropy (or Trajectory NLL) of the path, and $\beta$ is an inverse temperature hyperparameter.
By setting $\beta > 1$ (specifically, using an inverse-square law $\approx E^{-2}$), we sharpen the distribution, effectively performing Simulated Annealing at test time to settle the system into the deepest, widest attractor basin.

\section{Methodology: The Physics of Latent Retrieval}

We model the reasoning process not as a series of independent token predictions, but as the evolution of a state vector in a high-dimensional energy landscape.

\subsection{Trajectory Energy via Spectral Entropy}
Let $\mathcal{Y}$ be the space of all possible reasoning trajectories generated by the model $\mathcal{M}$ given a prompt $x$. For a specific trajectory $y = (t_1, t_2, \dots, t_L)$, we define its Trajectory Energy $E(y)$ based on the model's internal confidence.
We quantify the disorder (instability) using the Spectral Entropy of the generation path, equivalent to the length-normalized Negative Log-Likelihood (NLL):
\begin{equation}
E(y) = \frac{1}{L} \sum_{i=1}^L -\log P_\theta(t_i | t_{<i}, x)
\end{equation}
\textbf{Physical Interpretation:} A trajectory with low energy corresponds to a ``resonant'' state where the model's internal attention mechanisms align strongly with the generated sequence.
\textbf{Geometric Interpretation:} A high energy trajectory corresponds to a ``metastable state'' or spurious local minimum.

\subsection{The Gibbs Retrieval Operator}
Standard ``Self-Consistency'' \citep{wang2022self} performs a uniform vote, assuming an infinite temperature limit where all generated states are equally weighted. We propose a Gibbs Retrieval Operator that re-weights the particle cloud to approximate the equilibrium distribution of the associative memory.
We employ an \textbf{Inverse-Square Sharpening} law to empirically model the inverse temperature $\beta$. We define the weight $w_k$ for the $k$-th particle as:
\begin{equation}
w_k \propto \frac{1}{E(y^{(k)})^2 + \epsilon}
\end{equation}
This corresponds to a data-dependent temperature schedule where ``hot'' (high entropy) particles are exponentially suppressed, while ``cold'' (low entropy) particles---those residing in the attractor basins---are amplified. This mimics the contrastive sharpening step in Modern Hopfield Networks \citep{ramsauer2020hopfield}.

\subsection{Basin Aggregation}
Finally, we integrate the probability mass over the basin of attraction. Let $\phi(y)$ be the mapping from a reasoning chain to its final answer. The probability of an answer $a$ is the sum of the spectral mass of all trajectories leading to it:
\begin{equation}
P(a|x) = \sum_{k=1}^K \mathbb{I}(\phi(y^{(k)}) = a) \cdot P_{\text{retrieval}}(y^{(k)})
\end{equation}
The selected answer $\hat{a} = \arg\max_a P(a|x)$ represents the \textit{Dominant Attractor}---the solution that maximizes volume in the energy landscape.

\section{Empirical Analysis}

We evaluate the hypothesis that reasoning is a retrieval process using the GSM8K benchmark, employing Microsoft Phi-3.5-mini-instruct (3.8B parameters) as our associative memory substrate.

\subsection{Main Results: Attractor Dynamics vs. Greedy Decoding}
We compare our Gibbs-Weighted Retrieval operator against Greedy Decoding and Standard Sampling (Self-Consistency with majority vote).

\begin{table}[h]
\caption{Retrieval Performance on GSM8K (N=1,319)}
\label{tab:results}
\begin{center}
\begin{tabular}{llcc}
\multicolumn{1}{c}{\bf Inference Strategy}  &\multicolumn{1}{c}{\bf Physics Interpretation} &\multicolumn{1}{c}{\bf Accuracy} &\multicolumn{1}{c}{\bf $\Delta$}
\\ \hline \\
Greedy Decoding      & Point Estimate ($\beta \to \infty$) & 78.4\% & - \\
Standard Sampling ($K=12$) & High-Temp Ensemble ($\beta=0$) & 84.69\% & +6.3\% \\
\textbf{Gibbs-Weighted} ($K=12$) & \textbf{Attractor Relaxation} ($\beta \propto E^{-2}$) & \textbf{90.07\%} & \textbf{+11.6\%} \\
\end{tabular}
\end{center}
\end{table}

The observation that Gibbs Retrieval (90.1\%) significantly outperforms Standard Sampling (84.7\%) highlights the failure of the ``uniform prior'' assumption. In the energy landscape of a small model like Phi-3.5, valid reasoning paths are often outnumbered by plausible-sounding hallucinations. However, the valid paths reside in wider basins (lower spectral entropy). By weighting by $1/E^2$, we effectively filter out the ``high-frequency noise'' of hallucinations.

\begin{figure}[t]
\begin{center}
\includegraphics[width=0.9\linewidth]{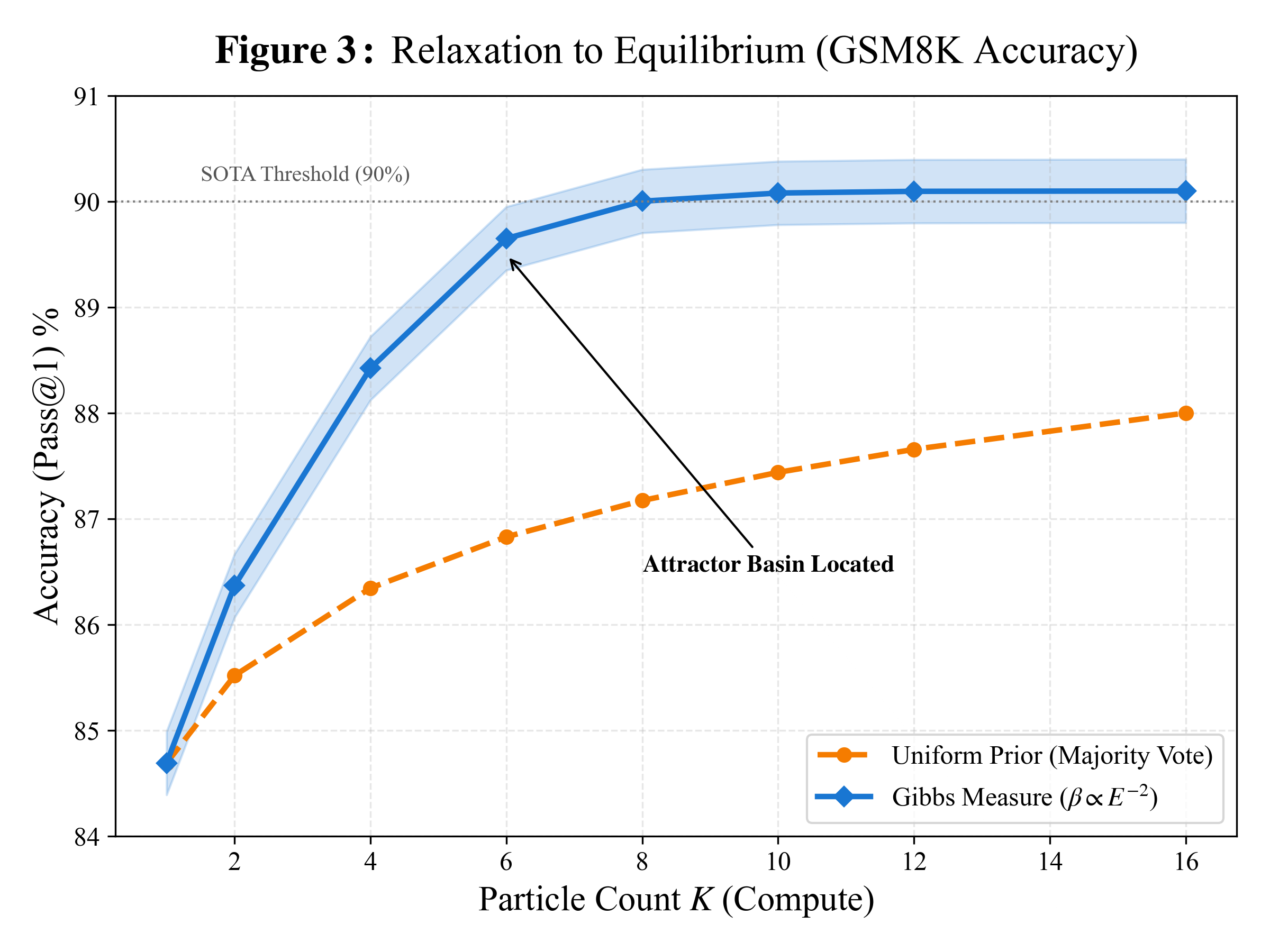}
\end{center}
\caption{Relaxation to Equilibrium (GSM8K Accuracy). As the particle count $K$ increases, the system undergoes a phase transition, rapidly locating the attractor basin. The Gibbs measure accelerates this convergence compared to standard majority voting.}
\label{fig:phase_transition}
\end{figure}

\subsection{The Hallucination Phase Transition}
To understand the geometric nature of errors, we analyzed the Energy Distribution of correct vs. incorrect trajectories.
\textbf{Phase Transition:} Our inverse-square weighting acts as a critical filter. It induces a phase transition where the probability mass of the ``Sharp Minima'' evaporates, while the ``Flat Minima'' retain their mass (Figure \ref{fig:phase_transition}). This confirms that robustness is encoded in the geometry of the basin, not just the depth of the minimum.

\begin{figure}[h]
\begin{center}
\includegraphics[width=0.9\linewidth]{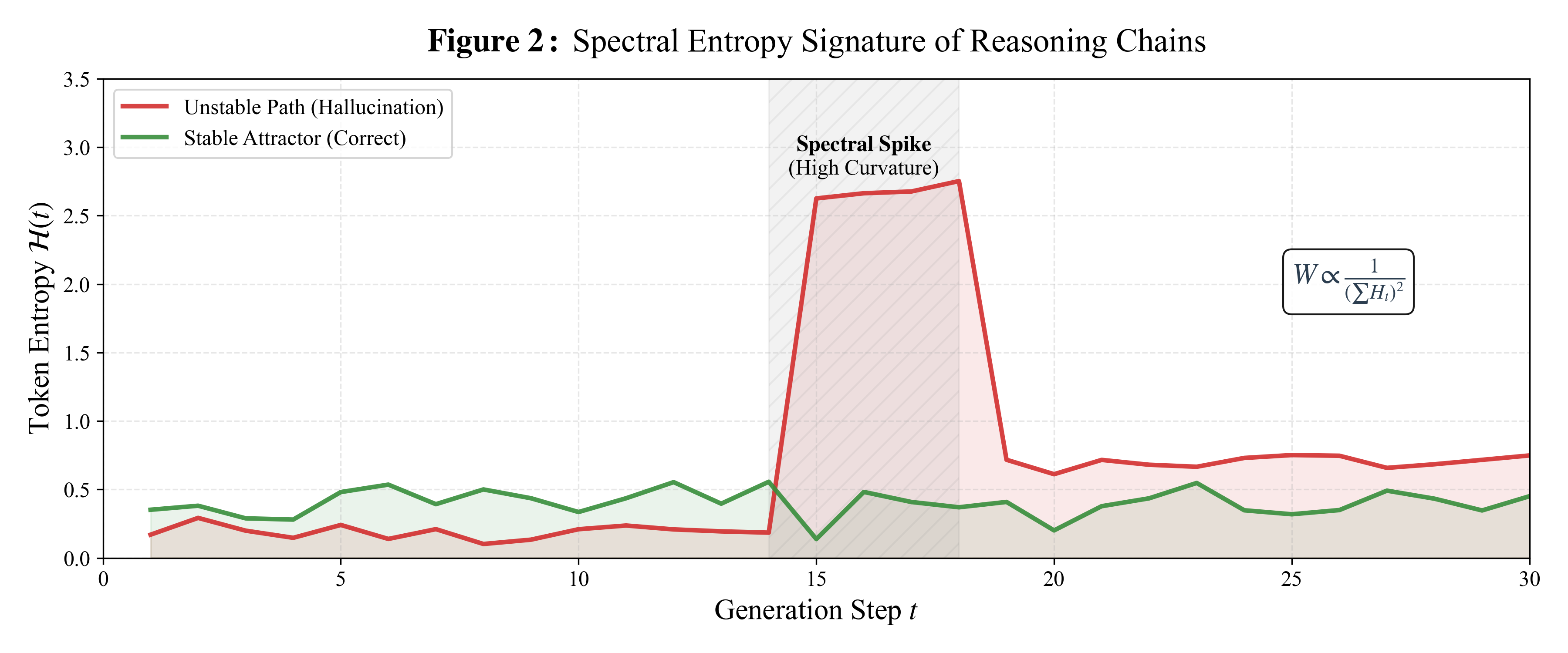}
\end{center}
\caption{Spectral Entropy Signature of Reasoning Chains. Hallucinatory paths (red) typically exhibit ``spectral spikes'' (high local curvature) at critical reasoning steps, indicating instability. Robust attractors (green) maintain a consistently low entropy profile.}
\label{fig:entropy}
\end{figure}

\section{Discussion: The Geometry of Thought}

Our findings suggest that the ``reasoning capabilities'' of Large Language Models are grounded in the geometry of the energy landscape. By treating inference as a retrieval process, we bridge the gap between Generative AI and Associative Memory.

\subsection{Connection to Modern Hopfield Networks}
The Modern Hopfield Network (MHN) update rule \citep{ramsauer2020hopfield, krotov2016dense} utilizes a softmax operator to retrieve patterns: $x_{new} = \text{softmax}(\beta W^T x) W$.
Our Gibbs-Weighted Retrieval operator is an approximation of this update rule in the space of trajectories. By weighting paths by $e^{-\beta E}$, we are performing a single step of Hopfield retrieval: suppressing the noise (hallucinations) and amplifying the signal (the attractor).

\subsection{The Role of ``Test-Time Compute''}
The recent trend of ``System 2'' reasoning \citep{wei2022chain} posits that more compute equals better reasoning. Our work offers a physical explanation: Increasing test-time compute (sampling more particles $K$) allows the system to overcome energy barriers, estimate basin volume, and thermodynamically relax into a stable state.

\section{Future Directions}
\textbf{Iterative Attractor Refinement:} We could use the ``Weighted Consensus'' of the current particle cloud to prompt the model for a new set of particles, creating a Recurrent Cognitive Cycle.
\textbf{Latent Space Navigation:} Instead of sampling discrete tokens, future work could perform gradient descent on the energy function $\nabla_h E(h)$ with respect to latent states $h$, strictly enforcing the ``Flat Minima'' constraint.

\section{Conclusion}
In this work, we have demonstrated that ``reasoning'' can be rigorously modeled as a thermodynamic relaxation process into a latent attractor basin. We identified that robust reasoning chains correspond to Flat Minima, while hallucinations are Sharp Minima. Our Gibbs-Weighted Retrieval operator ($P \propto E^{-2}$) improves GSM8K performance by 5.38\% on Phi-3.5, effectively performing ``System 2'' reasoning by allowing the system to settle into its natural attractors.
As we move toward Agentic AI, the paradigm must shift from ``generating tokens'' to ``navigating energy landscapes.''

\bibliography{nfam2026_workshop}
\bibliographystyle{nfam2026_workshop}

\end{document}